\documentclass[times, 10pt,review]{elsarticle}

\usepackage{amssymb}
\usepackage{amsmath}

\usepackage{multirow, xcolor, subfigure, booktabs, algorithm, algorithmic}

\journal{Pattern Recognition}

\begin{document}

\begin{frontmatter}



\title{FastFace: Fast-converging Scheduler for Large-scale Face Recognition Training with One GPU}


\author[JNU1]{Xueyuan Gong} 
\ead{xygong@jnu.edu.cn}
\author[JNU2]{Zhiquan Liu}
\ead{zqliu@vip.qq.com}
\author[UM]{Yain-Whar Si}
\ead{fstasp@um.edu.mo}
\author[MPU]{Xiaochen Yuan}
\ead{xcyuan@mpu.edu.mo}
\author[JNU3]{Ke Wang}
\ead{wangke@jnu.edu.cn}
\author[JNU1]{Xiaoxiang Liu}
\ead{tlxx@jnu.edu.cn}
\author[JNU1]{Cong Lin}
\ead{conglin@jnu.edu.cn}
\author[JNU1]{Xinyuan Zhang\corref{corresponding}}
\ead{zhangxy@jnu.edu.cn}
\cortext[corresponding]{Corresponding author}

\affiliation[JNU1]{
	organization={School of Intelligent Systems Science and Engineering, Jinan University},
	city={Zhuhai},
	country={China}
}

\affiliation[JNU2]{
	organization={College of Cyber Security, Jinan University},
	city={Guangzhou},
	country={China}
}

\affiliation[UM]{
	organization={Faculty of Science and Technology, University of Macau},
	city={Macau},
	country={China}
}

\affiliation[MPU]{
	organization={Faculty of Applied Sciences, Macau Polytechnic University},
	city={Macau},
	country={China}
}

\affiliation[JNU3]{
	organization={College of Information Science and Technology, Jinan University},
	city={Guangzhou},
	country={China}
}

\begin{abstract}
Computing power has evolved into a foundational and indispensable resource in the area of deep learning, particularly in tasks such as Face Recognition (FR) model training on large-scale datasets, where multiple GPUs are often a necessity. Recognizing this challenge, some FR methods have started exploring ways to compress the fully-connected layer in FR models. Unlike other approaches, our observations reveal that without prompt scheduling of the learning rate (LR) during FR model training, the loss curve tends to exhibit numerous stationary subsequences. To address this issue, we introduce a novel LR scheduler leveraging Exponential Moving Average (EMA) and Haar Convolutional Kernel (HCK) to eliminate stationary subsequences, resulting in a significant reduction in converging time. However, the proposed scheduler incurs a considerable computational overhead due to its time complexity. To overcome this limitation, we propose FastFace, a fast-converging scheduler with negligible time complexity, i.e. $\mathcal{O}(1)$ per iteration, during training. In practice, FastFace is able to accelerate FR model training to a quarter of its original time without sacrificing more than $1\%$ accuracy, making large-scale FR training feasible even with just one single GPU in terms of both time and space complexity. Extensive experiments validate the efficiency and effectiveness of FastFace. The code is publicly available at: https://github.com/amoonfana/FastFace
\end{abstract}




\end{frontmatter}

\section{Introduction}

Face recognition (FR) plays a crucial role in real-life applications. In recent years, the scale of FR training datasets has grown from $0.5$M (Million) images with $0.01$M IDs (Identities) in CASIA Webface~\cite{Yi:2014} to $42$M images with $2$M IDs in WebFace42M~\cite{Zhu:2021}, which indicates the training of deep FR models have entered the large-scale datasets era, called large-scale FR training.

Deep FR models have seen significant advancements due to large-scale training datasets, such as MegaFace~\cite{Shlizerman：2016}, MS1M~\cite{Guo:2016}, and WebFace42M~\cite{Zhu:2021}. Yet, the basic requirements for training deep FR models have also escalated, evolving from $1$ node (Personal computer or server) with $4\times$GPUs to $1$ node with $8\times$GPUs. The state-of-the-art FR approach, Partial FC (PFC)~\cite{An:2022}, utilizes up to $8$ nodes with $8\times$GPUs each, amounting to a total of $64\times$GPUs. This substantial demand for computing power has effectively barred many researchers from engaging in large-scale FR training. In fact, computing power has become a fundamental and critical resource in the deep learning area, making it nearly impossible to train a deep FR model without access to sufficient hardware resources.

\begin{figure}[!t]
	\centering
	\subfigure[The loss curve]{
		\includegraphics[width=0.47\columnwidth]{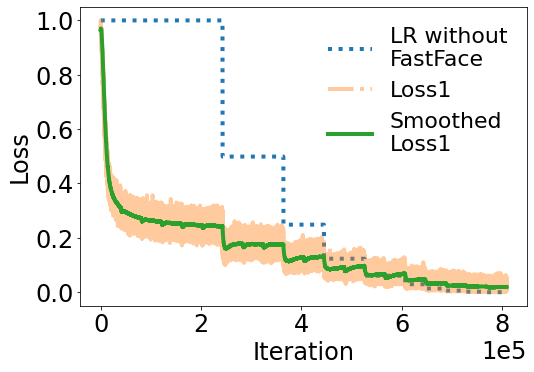}
	}
	\hfil
	\subfigure[The loss curve with FastFace]{
		\includegraphics[width=0.47\columnwidth]{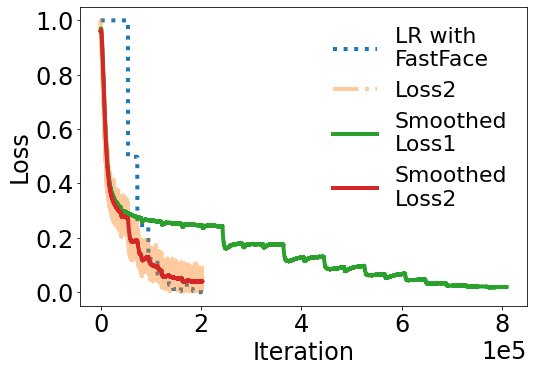}
	}
	\caption{The loss curve of training ResNet100 on MS1MV3. All curves are normalized to $[0,1]$ for putting in one figure. With FastFace, the loss after training $2e5$ steps is close to that after training $8e5$ steps without FastFace. Thus, only $1/4$ of its original training time is required, which enables us to train models with $1\times$GPU in a feasible time while preserving the accuracy.}
	\label{fg:stair_case}
\end{figure}

To address the challenge of computational resources, many cutting-edge FR approaches, e.g. PFC~\cite{An:2022}, Faster Face Classification (F$^{2}$C)~\cite{Wang:2022}, Dynamic Class Queue (DCQ)~\cite{BLi:2021}, and Virtual Fully-connected (Virtual FC)~\cite{PYLi:2021}, have already started to consider the problem of large-scale FR training. Nevertheless, those approaches primarily concentrate on reducing the space complexity of the fully-connected layer in the model. As a result, large-scale FR training with multiple GPUs remains a fundamental requirement to ensure the feasible training time. In response to this, we propose a method to accelerate the converging time of the model for large-scale FR training using $1\times$GPU.

Till now, the common practice for large-scale FR training is to train models for more than $20$ epochs using at least $4\times$GPUs. In addition, a learning rate (LR) scheduler is employed to stabilize the convergence of models. However, we observed that the loss curve would be shaped like a staircase while doing multiple step LR scheduling, as shown in Fig.~\ref{fg:stair_case}(a). This phenomenon indicates that the fast and accurate LR scheduling will help eliminate those stationary subsequences, in order to reduce the converging time of the model. Thus, the intuition is to schedule the LR promptly and accurately once the loss curve goes stationary, as shown in Fig.~\ref{fg:stair_case}(b). In this way, the deep FR model can converge more quickly without sacrificing much accuracy.

In this paper, we propose a simple yet effective scheduler, which leverages Exponential Moving Average (EMA) and Haar Convolutional Kernel (HCK). Nonetheless, the proposed scheduler incurs a considerable computational overhead. To overcome this limitation, we propose FastFace, a fast-converging scheduler which schedules the LR promptly and accurately, effectively eliminating long stationary subsequences in the loss curve, in order to accelerate the converging time of FR models. As a consequence, the large-scale FR training time with $1\times$GPU can be close to that with $8\times$GPUs. The contribution of this paper can be summarized as follows:

\begin{itemize}
	\item \textbf{Efficient.} The time complexity of FastFace is a negligible overhead, i.e. $\mathcal{O}(1)$ per iteration. In addition, FastFace enables the FR model to converge in $5$ epochs on $1\times$GPU, which is $1/4$ of its original converging time, i.e. $20$ epochs on $8\times$GPUs.
	\item \textbf{Effective.} FastFace sacrifices negligible accuracy while reducing training epochs from $20$ to $5$. Specifically, when training ResNet100 on WebFace12M, FastFace achieves an accuracy of $97.20$ after $5$ epochs on the dataset IJB-C, compared to $97.58$ without FastFace after $20$ epochs. Thus, we conclude the accuracy loss of FastFace is less than $1\%$.
	\item \textbf{Impactful.} FastFace demonstrates that large-scale FR training with $1\times$GPU is not a mission impossible. It opens the door for many researchers to engage in large-scale FR training without requiring massive hardware resources.
\end{itemize}

\section{Related Work}

In this section, we mainly reviewed the FR approaches relating to PFC~\cite{An:2022}, since it is one of the state-of-the-art approaches aiming at large-scale FR training.

Back in 2017, SphereFace~\cite{Liu:2017} had already started to train FR models with $4\times$GPUs, pioneering the approach of mapping the face image onto a hyper-sphere in the feature space. CosFace~\cite{HWang:2018} proposed to normalize both the feature vector and the class vector in order to stabilize the training process, utilizing $8\times$GPUs for FR model training. UniformFace~\cite{Duan:2019} introduced learning equidistributed feature vectors in order to maximize feature space utilization, employing $4\times$GPUs. ArcFace~\cite{Deng1:2019} proposed an additive angular margin in order to maximize the inter-class separation and the intra-class compactness. It trained FR models with $4\times$GPUs. CurricularFace~\cite{Huang:2020} embedded the idea of curriculum learning into the loss function, in order to address easy samples in the early training stage and hard ones in the later stage. It utilized $4\times$GPUs in the FR model training. MagFace~\cite{Meng:2021} linked image quality to the magnitude of their feature vectors, in order to improve the generalization ability of the model. It leveraged $8\times$GPUs in the FR model training. ElasticFace~\cite{Boutros:2022} proposed to allow flexibility in the push for inter-class separability, in order to give the decision boundary chances to extract and retract to allow space for flexible class separability learning, training with $4\times$GPUs. There also exists many other methods using multiple GPUs, e.g. CoReFace~\cite{Song:2024} with $4\times$GPUs, Hybrid tOken Transformer (HOTformer)~\cite{Su:2023} with $8\times$GPUs, etc. However, there are still some FR approaches where the specific hardware details are not provided, such as Large-Margin Softmax (L-Softmax) loss~\cite{Liu:2016}, Additive Margin Softmax (AM-Softmax) loss~\cite{FWang:2018}, Circle loss~\cite{Sun:2020}, and AdaFace~\cite{Kim:2022}. To sum up, almost all FR approaches rely on multiple GPUs for training, indicating that this is a common practice in the field, with few exceptions where hardware information is not available.

Recently, some FR approaches have begun addressing the challenges of large-scale FR training. Dynamic Class Queue (DCQ)~\cite{BLi:2021} proposed dynamically selecting and generating class vectors instead of using them all during training. It adopted $8\times$GPUs to train FR models. Similar to DCQ, Faster Face Classification (F$^{2}$C)~\cite{Wang:2022} introduced Dynamic Class Pool (DCP) to dynamically store and update class vectors. It employed $8\times$GPUs for the FR model training. Instead of maintaining a queue or a pool, Partial Fully-Connected (PFC)~\cite{An:2022} suggested randomly selecting negative class vectors. It leveraged at least $8\times$GPUs for the FR model training. Virtual Fully-connected (Virtual FC)~\cite{PYLi:2021} shares a similar goal with our approach, aiming to enable large-scale FR training with limited hardware resources. However, while Virtual FC focuses on reducing the space and time complexity of the FC layer, our approach targets reducing the converging time. These two strategies are different and non-conflicting. In addition, Some researchers utilized Knowledge Distillation (KD) to compress the model, in order to acquire a more efficient inference time while preserving the accuracy, such as Teacher Guided Neural Architecture Search~\cite{Wang:2021} and Grouped-KD~\cite{Zhao:2023}

\section{Problem Statement}

In this section, we first revisit the large-scale FR training. After that, the problem in this paper is discussed.

A face image is defined as $\mathbf{x}\in\mathbb{R}^{N\times N\times 3}$, where $N$ is the size of the image and $3$ is the channel of it. Thus, a deep FR model is defined as $\mathbf{f}=\mathcal{F}(\mathbf{x})$, where $\mathbf{f}\in\mathbb{R}^{D}$ is the feature vector extracted from $\mathbf{x}$. In order to train the model, the equation of combined margin loss function used in ArcFace and Partial FC is reviewed as follows:

\begin{equation}
	\label{eq:cml}
	\begin{aligned}
		&\mathcal{L}=-\frac{1}{B}\sum_{i=1}^{B}\log\frac{e^{s\mathcal{M}(\theta_{y_{i},i})}}{e^{s\mathcal{M}(\theta_{y_{i},i})}+\sum_{j=1,j\notin y_{i}}^{C}e^{s\cos\theta_{j,i}}}\\
		&\mathcal{M}(\theta_{y_{i},i})=\cos(m_{1}\theta_{y_{i},i}+m_{2})-m_{3}
	\end{aligned}
\end{equation}
where $B$ denotes the batch size, $C$ is the number of classes, $\mathcal{M}$ means the marginal function, $m_{1}$ stands for the multiplicative angular margin, $m_{2}$ represents the additive angular margin, $m_{3}$ denotes the additive cosine margin, $s$ is the scale factor controlling the radius of the hyper-sphere in feature space, $\theta_{j,i}$ represents the angle between the class vector $\mathbf{w}_{j}$ and the feature vector $\mathbf{f}_{i}$, $\mathbf{w}_{j}\in\mathbb{R}^{D}$ stands for the $j$-th class vector, and $\mathbf{f}_{i}$ denotes the feature vector of the $i$-th image belonging to the $y_{i}$-th class. Thus, $\theta_{y_{i},i}$ can be calculated by the equation $\theta_{y_{i},i}=\arccos(\mathbf{w}_{y_{i}}\mathbf{f}_{i}/\Vert\mathbf{w}_{y_{i}}\Vert\Vert\mathbf{f}_{i}\Vert)$, and $\mathcal{M}(\theta_{y_{i},i})$ can be computed afterwards. Note that $m_{1}$ will change the frequency of the $\cos(\cdot)$ function and destroy its monotonicity. Thus, we set $m_{1}$ as $1$ in this paper and thus it is treated as a constant.

To this end, we are able to train the model $\mathcal{F}$. Specifically, $\mathcal{F}$ is trained for $E$ epochs, where each epoch contains $U$ iterations. Thus, $\mathcal{F}$ is updated $T=EU$ times in total, where the model and the loss at the $t$-th iteration are denoted as $\mathcal{F}_{t}$ and $\mathcal{L}_{t}$ respectively. Formally, all $\mathcal{F}_{t}$ forms a sequence (time-series) $\mathbf{F}=\{\mathcal{F}_{0}, \mathcal{F}_{1}, \dots, \mathcal{F}_{T}\}$. Similarly, all $\mathcal{L}_{t}$ forms another sequence $\mathbf{L}=\{\mathcal{L}_{0}, \mathcal{L}_{1}, \dots, \mathcal{L}_{T}\}$. Hence, we wish to identify a sequence of the learning rate (LR) $\mathbf{\Gamma}=\{\gamma_{0}, \gamma_{1}, \dots, \gamma_{T}\}$, in order to minimize the final loss $\mathcal{L}_{T}$ with as less $T$ as possible.

In practice, the LR $\gamma$ is scheduled manually in the training process, which means the scheduling strategy is a hyper-parameter. The majority of FR approaches select Multiple Step LR (MultiStepLR) scheduler, where $\gamma_{t}$ is reduced by a decay factor $\delta$ after certain training iterations. The new LR $\gamma_{t+1}$ can be calculated by $\gamma_{t+1}=\gamma_{t}/\delta$. However, as already shown in Fig.~\ref{fg:stair_case}(a), we discover that MultiStepLR always brings many stationary subsequences in the loss curve. It is potential to accelerate the converging time of the model by eliminating those stationary subsequences. Therefore, FastFace is proposed in the paper.

\section{The Proposed Approach}

In this section, FastFace is introduced from three subsections. The first subsection presents Exponential Moving Average (EMA) to smooth the loss curve. The second subsection demonstrates Haar Convolutional Kernel (HCK) to detect stationary subsequences, and combines HCK with EMA to propose a naive method called EMA+HCK. The third subsection reduces the computational overhead of EMA+HCK and proposes FastFace.

\subsection{Exponential Moving Average}

\begin{figure}[!t]
	\centering
	\subfigure[The loss curve $\mathbf{L}$ before smoothing]{
		\includegraphics[width=0.47\columnwidth]{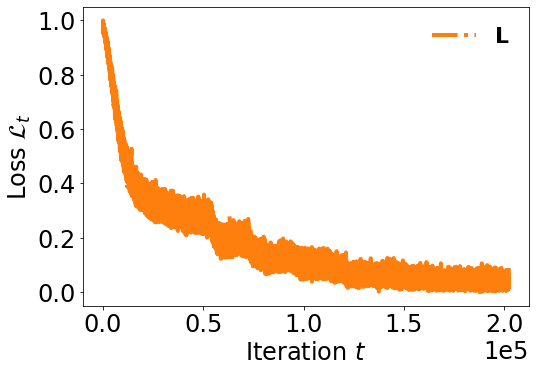}
	}
	\hfil
	\subfigure[The loss curve $\mathbf{L}^{E}$ after smoothing]{
		\includegraphics[width=0.47\columnwidth]{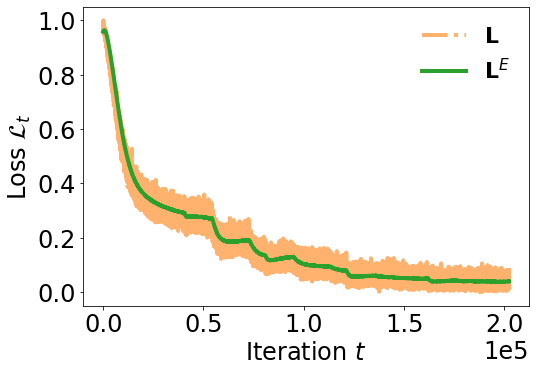}
	}
	\caption{Smoothing the loss curve by EMA. As shown in (a), the original loss curve is hard to be analyzed since it contains too much noise. Thus, as shown in (b), we adopt EMA to smooth it. Note all curves are normalized}
	\label{fg:ema}
\end{figure}

Given a loss sequence $\mathbf{L}=\{\mathcal{L}_{0}, \mathcal{L}_{1}, \dots, \mathcal{L}_{T}\}$, it is hard to detect the stationary subsequences as it contains too many noises, which are caused by the optimization algorithm, i.e. Mini-batch Gradient Descent. More specifically, Batch Gradient Descent can generate a more smoothed loss sequence $\mathbf{L}$, since it considers all samples (face images) in every iteration. Thus, the loss function $\mathcal{L}$ remains unchanged. On the contrary, Mini-batch Gradient Descent considers distinct $B$ samples in every iteration. Accordingly, the loss function $\mathcal{L}$ will always change in each iteration. This leads to the result that the generated loss sequence $\mathbf{L}$ is unstable, which contains too many noises, as shown in Fig.~\ref{fg:ema}(a). Therefore, Exponential Moving Average (EMA) is selected to smooth the loss sequence $\mathbf{L}$, which filters noises to help the detection of stationary subsequences. The expected $\mathbf{L}$ after smoothing is shown in Fig.~\ref{fg:ema}(b).

There are two reasons why we adopt EMA: 1) The loss sequence $\mathbf{L}$ is generated on-the-fly in the training process. EMA supports real-time updating, which is just compatible with our problem; 2) The time complexity of EMA in each iteration is only $\mathcal{O}(1)$, which indicates that it occupies negligible computations. The equation of EMA is given in Eq. \eqref{eq:ema}.

\begin{equation}
	\label{eq:ema}
	\mathcal{L}_{t}^{E}=
	\begin{cases}
		\mathcal{L}_{t},\hfill t=0\\
		\alpha\mathcal{L}_{t}+(1-\alpha)\mathcal{L}_{t-1}^{E},\hfill t>0
	\end{cases}
\end{equation}
where $\alpha\in[0,1]$ indicates the importance of the current loss $\mathcal{L}_{t}$ compared to the average of previous losses $\mathcal{L}_{t-1}^{E}$ ($t\in\{0,1,\dots,T\}$). Initially, when $t=0$, $\mathcal{L}_{0}^{E}=\mathcal{L}_{0}$ directly. After that, $\mathcal{L}_{t}^{E}$ is calculated by the current loss $\mathcal{L}_{t}$ and the average of previous losses $\mathcal{L}_{t-1}^{E}$. In this paper, $\alpha$ is set as $0.001$.

Overall, EMA generates a smoothed loss sequence  $\mathbf{L}^{E}=\{\mathcal{L}_{0}^{E}, \mathcal{L}_{1}^{E}, \dots, \mathcal{L}_{T}^{E}\}$ on-the-fly while generating the loss sequence $\mathbf{L}$, in order to support the detection of stationary subsequences.

\subsection{Haar Convolutional Kernel}

With EMA calculating $\mathbf{L}^{E}$ on-the-fly, we are able to detect stationary subsequences in real-time. Formally, given a smoothed loss sequence $\mathbf{L}^{E}=\{\mathcal{L}_{0}^{E}, \mathcal{L}_{1}^{E}, \dots, \mathcal{L}_{T}^{E}\}$, we wish to identify if the subsequence of $\mathbf{L}^{E}$ starting from $S$ and ending at $T$, denoted $\mathbf{L}_{S,T}^{E}=\{\mathcal{L}_{S}^{E}, \mathcal{L}_{S+1}^{E}, \dots, \mathcal{L}_{T}^{E}\}$, keeps stationary statistically, where $0\leq S<T$. Note $\mathcal{L}_{T}^{E}$ is always the last value in $\mathbf{L}^{E}$ and it increases continuously, since $\mathbf{L}^{E}$ is updating in real-time.

In this paper, we adopt Haar Convolutional Kernel (HCK) in the Haar-like Feature Descriptors to detect stationary subsequences. HCK is defined as $\mathcal{H}_{2s}$, where $s$ represents the half size of it, the first half values of $\mathcal{H}_{2s}$ from $1$ to $s$ are $-1$, the second half values of $\mathcal{H}_{2s}$ from $s+1$ to $2s$ are $1$. For example, $\mathcal{H}_{4}=\{-1,-1,1,1\}$. Thus, performing a convolutional operation on $\mathbf{L}^{E}$ using $\mathcal{H}_{2s}$, denoted $\mathcal{H}_{2s}*\mathbf{L}^{E}$, indicates calculating the difference between the sum of the first half values in $\mathbf{L}_{S,T}^{E}$ and the sum of the second half values in $\mathbf{L}_{S,T}^{E}$, where the size of $\mathbf{L}_{S,T}^{E}$ is $2s$. In this way, a stationary $\mathbf{L}_{S,T}^{E}$ will get a small difference value, while a $\mathbf{L}_{S,T}^{E}$ with a decreasing trend will get a relatively larger difference value. Hence, we are now able to distinguish the decreasing curve and the stationary curve. This straightforward approach combining EMA and HCK is named as EMA+HCK.

Nevertheless, the large size of $\mathcal{H}_{2s}$ in EMA+HCK will cause two drawbacks: 1) The time complexity of computing a single step convolution between $\mathcal{H}_{2s}$ and $\mathbf{L}_{S,T}^{E}$ is $\mathcal{O}(2s)$. It is inefficient to compute a long $\mathcal{H}_{2s}$ in every training iteration. For example, if $s=5000$, a $10$K times multiplication and a $10$K times summation are needed in each iteration. In large-scale FR training, the total iterations $T$ is commonly more than $500$K, which results in an extra $10$B (Billion) times computations in total; 2) A long $\mathcal{H}_{2s}$ requires a delay in detection. For instance, if $s=5000$, then we have to wait until $\mathbf{L}^{E}$ accumulates $10$K values before starting the detection, since the convolutional operation requires the size of $\mathbf{L}^{E}$ to be longer than $\mathcal{H}_{2s}$. However, we wish the stationary subsequence detection could start as soon as possible. Thus, we propose FastFace to reduce the computational overhead and the detection delay.

\subsection{FastFace}

\begin{figure}[!t]
	\centering
	\subfigure[The curve of $\mathbf{D}$]{
		\includegraphics[width=0.47\columnwidth]{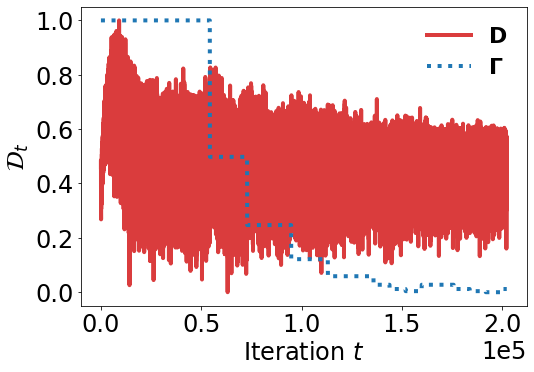}
	}
	\hfil
	\subfigure[The curve of $\mathbf{D}^{E}$]{
		\includegraphics[width=0.47\columnwidth]{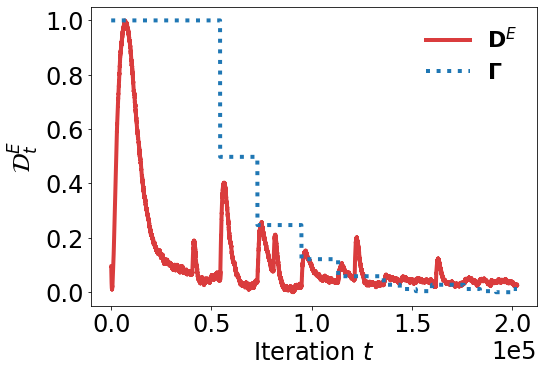}
	}
	\caption{Examples of $\mathbf{D}$ and $\mathbf{D}^{E}$. As shown in (a), it is hard to find a relationship between $\mathbf{L}$ and $\mathbf{D}$. By contrast, as shown in (b), $\mathbf{D}$ is smoothed by EMA and denoted as $\mathbf{D}^{E}$. It is clear to observe a spike on $\mathbf{D}^{E}$ every time when there is a decline on $\mathbf{L}$. Note all curves are normalized}
	\label{fg:dif_ema}
\end{figure}

When $s$ is set to $1$, it requires negligible computations and the shortest delay. Moreover, the convolutional operation $\mathcal{H}_{2}*\mathbf{L}^{E}$ degrades into calculating the difference $\mathcal{D}_{t}$ of $\mathcal{L}_{t-1}^{E}$ and $\mathcal{L}_{t}^{E}$. We denote the sequence of $\mathcal{D}_{t}$ as $\mathbf{D}=\{\mathcal{D}_{1},\mathcal{D}_{2},\dots,\mathcal{D}_{T}\}$ which can be calculated on-the-fly. The equation of it is given below:

\begin{equation}
	\label{eq:diff}
	\mathcal{D}_{t} = \mathcal{L}_{t-1}^{E} - \mathcal{L}_{t}^{E}
\end{equation}
where $t\in\{1,2,\dots,T\}$. Nonetheless, considering only two values in each step leads to a result that $\mathbf{D}$ is sensitive to noises. As shown in Fig.~\ref{fg:dif_ema}(a), it looks like a meaningless sequence, which is hard to use.

Therefore, we leverage EMA the second time, which aims at smoothing $\mathbf{D}$. The smoothed $\mathbf{D}$ is denoted as $\mathbf{D}^{E}=\{\mathcal{D}_{1}^{E}, \mathcal{D}_{2}^{E}, \dots, \mathcal{D}_{T}^{E}\}$. The equation of calculating $\mathbf{D}^{E}$ is shown as follows:

\begin{equation}
	\label{eq:dema}
	\mathcal{D}_{t}^{E}=
	\begin{cases}
		\mathcal{D}_{t},\hfill t=1\\
		\beta\mathcal{D}_{t}+(1-\beta)\mathcal{D}_{t-1}^{E},\hfill t>1
	\end{cases}
\end{equation}
where $\beta\in[0,1]$ indicates the importance of the current difference $\mathcal{D}_{t}$ compared to the average of previous differences $\mathcal{D}_{t-1}^{E}$. In this paper, $\beta$ is set as $0.001$.

Consequently, $\mathbf{D}^{E}$ works as a signal identifying stationary subsequences, as shown in Fig.~\ref{fg:dif_ema}(b). In detail, we are able to observe a spike on $\mathbf{D}^{E}$ every time when halving the LR $\gamma$, resulting in a decline on $\mathbf{L}$, which indicates a rapid drop in the loss curve. Thus, the subsequence with small values before the spike on $\mathbf{D}^{E}$ represents the stationary part on $\mathbf{L}$. We conclude that FastFace can schedule $\gamma_{t}$ once identifying a consecutive subsequence of small $\mathcal{D}_{t}^{E}$.

To this end, we propose to combine Eq. \eqref{eq:ema} (EMA), Eq. \eqref{eq:diff} (HCK with $s=1$), and Eq. \eqref{eq:dema} (EMA of $\mathbf{D}$) together. Thus, the calculation and storage for $\mathbf{L}^{E}$ and $\mathbf{D}$ are omitted. The combined approach is a novel fast-converging scheduler, called FastFace. FastFace is able to detect the stationary subsequences on-the-fly in the training process. Its time and space complexity are both $\mathcal{O}(1)$ in each iteration.

First, with Eq.~\eqref{eq:diff} and Eq.~\eqref{eq:ema}, we are able to get a variant of the equation for calculating $\mathcal{D}_{t}$, as given in Eq.~\eqref{eq:vdiff}.

\begin{equation}
	\label{eq:vdiff}
	\begin{aligned}
		\mathcal{D}_{t}&=\mathcal{L}_{t-1}^{E}-\mathcal{L}_{t}^{E}\\
		&=\mathcal{L}_{t-1}^{E}-\alpha\mathcal{L}_{t}-(1-\alpha)\mathcal{L}_{t-1}^{E}\\
		&=\alpha\mathcal{L}_{t-1}^{E}-\alpha\mathcal{L}_{t}\\
	\end{aligned}
\end{equation}

Next, it is easy to have $\mathcal{D}_{t-1}=\alpha\mathcal{L}_{t-2}^{E}-\alpha\mathcal{L}_{t-1}$ from Eq.~\eqref{eq:vdiff}, where $\mathcal{D}_{t-1}$ just means one iteration before $\mathcal{D}_{t}$. After performing $\mathcal{D}_{t-1}-\mathcal{D}_{t}$, we have:

\begin{equation}
	\label{eq:d2-d1}
	\begin{aligned}
		\mathcal{D}_{t-1}-\mathcal{D}_{t}&=\alpha\mathcal{L}_{t-2}^{E}-\alpha\mathcal{L}_{t-1}-\alpha\mathcal{L}_{t-1}^{E}+\alpha\mathcal{L}_{t}\\
		&=\alpha(\mathcal{L}_{t-2}^{E}-\mathcal{L}_{t-1}^{E})+\alpha(\mathcal{L}_{t}-\mathcal{L}_{t-1})\\
		&=\alpha\mathcal{D}_{t-1}+\alpha(\mathcal{L}_{t}-\mathcal{L}_{t-1})
	\end{aligned}
\end{equation}

From Eq.~\eqref{eq:vdiff}, we know $\mathcal{D}_{1}=\alpha\mathcal{L}_{0}^{E}-\alpha\mathcal{L}_{1}$. Then, since $\mathcal{L}_{0}^{E}=\mathcal{L}_{0}$ in Eq.~\eqref{eq:ema}, we have $\mathcal{D}_{1}=\alpha(\mathcal{L}_{0}-\mathcal{L}_{1})$. At last, we are able to get the final equation of $\mathcal{D}_{t}$ from Eq.~\eqref{eq:d2-d1}, which is shown as follows:

\begin{equation}
	\label{eq:diff2}
	\mathcal{D}_{t} =
	\begin{cases}
		\alpha(\mathcal{L}_{t-1}-\mathcal{L}_{t}),\hfill t=1\\
		(1-\alpha)\mathcal{D}_{t-1}+\alpha(\mathcal{L}_{t-1}-\mathcal{L}_{t}),\hfill t>1
	\end{cases}
\end{equation}

From Eq.~\eqref{eq:dema}, we have the following two equations:

\begin{equation}
	\label{eq:d_t}
	D_{t}=\frac{\mathcal{D}_{t}^{E}-(1-\beta)\mathcal{D}_{t-1}^{E}}{\beta}
\end{equation}
\begin{equation}
	\label{eq:d_t-1}
	D_{t-1}=\frac{\mathcal{D}_{t-1}^{E}-(1-\beta)\mathcal{D}_{t-2}^{E}}{\beta}
\end{equation}

Substituting $D_{t}$ and $D_{t-1}$ in Eq. \eqref{eq:diff2} by Eq. \eqref{eq:d_t} and Eq. \eqref{eq:d_t-1}, we then have:

\begin{equation}
	\label{eq:de_t1}
		\frac{\mathcal{D}_{t}^{E}-(1-\beta)\mathcal{D}_{t-1}^{E}}{\beta}=(1-\alpha)\frac{\mathcal{D}_{t-1}^{E}-(1-\beta)\mathcal{D}_{t-2}^{E}}{\beta}\\+\alpha(\mathcal{L}_{t-1}-\mathcal{L}_{t})\\
\end{equation}

\begin{equation}
	\label{eq:de_t2}
	\mathcal{D}_{t}^{E}=(1-\alpha)[\mathcal{D}_{t-1}^{E}-(1-\beta)\mathcal{D}_{t-2}^{E}]\\+(1-\beta)\mathcal{D}_{t-1}^{E}+\alpha\beta(\mathcal{L}_{t-1}-\mathcal{L}_{t})\\
\end{equation}

\begin{equation}
	\label{eq:de_t3}
	\mathcal{D}_{t}^{E}=(1-\alpha+1-\beta)\mathcal{D}_{t-1}^{E}\\-(1-\alpha)(1-\beta)\mathcal{D}_{t-2}^{E}\\+\alpha\beta(\mathcal{L}_{t-1}-\mathcal{L}_{t})
\end{equation}

Denoting $\omega_{1}=(1-\alpha)+(1-\beta)$, $\omega_{2}=(1-\alpha)(1-\beta)$, and $\omega_{3}=\alpha\beta$. Thus, when $t>1$, as shown in Eq. (8), $\mathcal{D}_{t}^{E}=\omega_{1}\mathcal{D}_{t-1}^{E}-\omega_{2}\mathcal{D}_{t-2}^{E}+\omega_{3}(\mathcal{L}_{t-1}-\mathcal{L}_{t})$. Note that we omit the derivation of $\mathcal{D}_{t}^{E}$ when $t=0$ and $t=1$, as it is straight-forward to get. Finally, the formula of FastFace is concluded as:

\begin{equation}
	\label{eq:dema2}
	\mathcal{D}_{t}^{E}=
	\begin{cases}
		0, \hfill t=0\\
		\alpha(\mathcal{L}_{t-1}-\mathcal{L}_{t}),\hfill t=1\\
		\omega_{1}\mathcal{D}_{t-1}^{E}-\omega_{2}\mathcal{D}_{t-2}^{E}+\omega_{3}(\mathcal{L}_{t-1}-\mathcal{L}_{t}),\hfill t>1
	\end{cases}
\end{equation}
where $\omega_{1}=(1-\alpha)+(1-\beta)$, $\omega_{2}=(1-\alpha)(1-\beta)$, $\omega_{3}=\alpha\beta$. Note that $\omega_{1}$, $\omega_{2}$, and $\omega_{3}$ are able to be pre-computed in order to reduce computational overhead.

To this end, calculating $\mathcal{D}_{t}^{E}$ by Eq.~\eqref{eq:dema2} does not include $\mathcal{L}_{t}^{E}$ and $\mathcal{D}_{t}$ anymore. In addition, it has three advantages: 1) $\mathcal{D}_{t}^{E}$ can be calculated on-the-fly. It requires only $1$ step delay, i.e. when $t=0$; 2) Its time complexity is $\mathcal{O}(1)$ in each iteration; 3) Its space complexity is also $\mathcal{O}(1)$, since it only needs to store $\mathcal{D}_{t-1}^{E}$ and $\mathcal{D}_{t-2}^{E}$ in $\mathbf{D}$, and $\mathcal{L}_{t-1}$ in $\mathbf{L}$, instead of the whole sequences.

After calculating $\mathcal{D}_{t}^{E}$ by Eq.~\eqref{eq:dema2}, we set a threshold $\lambda$ in order to determine if $\mathcal{D}_{t}^{E}$ is small enough, which indicates the loss sequence $\mathbf{L}$ keeps stationary. In addition, a tolerance $\tau$ is defined in case that FastFace is too sensitive to $\lambda$. Specifically, FastFace will reduce the learning rate $\gamma_{t}$ if $\mathcal{D}_{t}^{E}$ keeps lower than $\lambda$, i.e. $\mathcal{D}_{t}^{E}<\lambda$, for $\tau$ iterations. After extensive experiments, we empirically set $\lambda=5e-5$ and $\tau=0.05T$. The algorithm of FastFace is shown in Alg. \ref{alg:ff}.

\begin{algorithm}[tb!]
	\caption{The algorithm of FastFace}
	\label{alg:ff}
	\textbf{Input}: Previous learning rate $\gamma_{t-1}$, and current loss $\mathcal{L}_{t}$\\
	\textbf{Output}: Current learning rate $\gamma_{t}$\\
	\textbf{Initialization}: $\mathcal{D}_{t-1}^{E}\gets0$, $\mathcal{D}_{t-2}^{E}\gets0$, previous loss $\mathcal{L}_{t-1}\gets0$, iterations count $t\gets1$, tolerance count $c\gets0$, threshold $\lambda\gets5e-5$, tolerance $\tau\gets0.05T$,  $\alpha\gets0.001$, $\beta\gets0.001$, $\omega_{1}\gets(1-\alpha)+(1-\beta)$, $\omega_{2}\gets(1-\alpha)(1-\beta)$, $\omega_{3}\gets\alpha\beta$
	\begin{algorithmic}[1] 
		\IF {$t=1$}
		\STATE $\mathcal{D}_{t}^{E}\gets\alpha(\mathcal{L}_{t-1}-\mathcal{L}_{t})$\hfill$\triangleright$ Eq. \eqref{eq:dema2}
		\ELSE
		\STATE $\mathcal{D}_{t}^{E}\gets\omega_{1}\mathcal{D}_{t-1}^{E}-\omega_{2}\mathcal{D}_{t-2}^{E}+\omega_{3}(\mathcal{L}_{t-1}-\mathcal{L}_{t})$\hfill$\triangleright$ Eq. \eqref{eq:dema2}
		\ENDIF
		\IF{$\mathcal{D}_{t}^{E}<\lambda$}
		\IF{$c<\tau$}
		\STATE $\gamma_{t}\gets \gamma_{t-1}$
		\STATE $c\gets c+1$
		\ELSE
		\STATE $\gamma_{t}\gets \gamma_{t-1}/2$
		\STATE $c\gets 0$
		\ENDIF
		\ENDIF
		\STATE $\mathcal{D}_{t-2}^{E}\gets\mathcal{D}_{t-1}^{E}$
		\STATE $\mathcal{D}_{t-1}^{E}\gets\mathcal{D}_{t}^{E}$
		\STATE $\mathcal{L}_{t-1}\gets\mathcal{L}_{t}$
		\STATE $t\gets t+1$
		\STATE \textbf{return} $\gamma_{t}$
	\end{algorithmic}
\end{algorithm}

\section{Experiments}

\subsection{Implementation Details}

\paragraph{\textbf{Datasets}} In this paper, we separately employ MS1MV2~\cite{Deng1:2019}, MS1MV3~\cite{Deng2:2019}, WebFace4M, WebFace8M, and WebFace12M~\cite{Zhu:2021} as the training sets. MS1MV2 is also called MS1M-ArcFace, which has more than $5.8$M images with $0.085$M identities. MS1MV3 is also called MS1M-RetinaFace, which contains more than $5$M images with $0.09$M IDs. WebFace42M~\cite{Zhu:2021} contains $42$M images with $2$M IDs. From WebFace42M, we randomly selected $10\%$, $20\%$, and $30\%$ of the samples to create WebFace4M, WebFace8M, and WebFace12M respectively. Hence, WebFace12M can have more than $12$M images with $0.6$M IDs in total.

Many public test sets are adopted to evaluate the performance of FastFace, including LFW~\cite{Huang:2007}, CFP-FP~\cite{Sengupta:2016}, AgeDB-30~\cite{Moschoglou:2017}, IJB-B~\cite{Whitelam:2017}, and IJB-C~\cite{Maze:2018}.



\paragraph{\textbf{Experimental Settings}} The experiments in the paper are implemented on a computer equipped with an Intel Core i9-13900K 3.00 GHz CPU, 32 GB RAM, and an NVIDIA GeForce RTX 4090 GPU. The operating system is Windows 11. In addition, the development environment is Anaconda 23.7.2 with Python 3.11.4 and Pytorch 2.0.1. Following the settings in recent papers ~\cite{An:2022,Deng1:2019}, input images are cropped and centered to the size of $112\times112\times3$. Each pixel in the image is normalized by subtracting $127.5$ and then being divided by $128$, which is mapped from $[0,255]$ to $[-1,1]$. Images are randomly flipped for data augmentation. Customized ResNet50 and ResNet100~\cite{An:2022,He:2016} are employed as the backbones. The ResNet model outputs the feature vector of dimension $512$. After that, feature vectors are input into the combined margin loss. For training set MS1MV3, $m_{1}=1$, $m_{2}=0.5$ and $m_{3}=0$. For training sets WebFace4M, WebFace8M, and WebFace12M, $m_{1}=1$, $m_{2}=0$ and $m_{3}=0.4$. The scale factor $s$ is set as $64$. In addition, the ResNet model is trained for $5$ epochs on different training sets with only one single GeForce RTX 4090 GPU. The batch size is set as $128$ samples. Stochastic Gradient Descent (SGD) with momentum $0.9$ and weight decay $5e-4$ is selected as the optimizer. The learning rate is set as $0.2$, the decay factor $\delta=2$, and it is automatically scheduled by FastFace. For EMA, $\alpha$ and $\beta$ are both set as $0.001$. The hyper-parameters in FastFace are set as $\lambda=1e-5$ and $\tau=0.05$.

For testing, two feature vectors extracted from the original image and its horizontal-flipped one are summed together as the final feature vector. In addition, True Accept Rate (TAR) @ False Positive Rate (FAR) $=1e-4$ is reported on IJB-B and IJB-C, denoted as TAR@FAR$=1e-4$.

\subsection{Hyperparameter Study}

\begin{figure}[!t]
	\centering
	\subfigure[The heatmap of $\gamma_{0}$ and $\delta$]{
		\includegraphics[width=0.47\columnwidth]{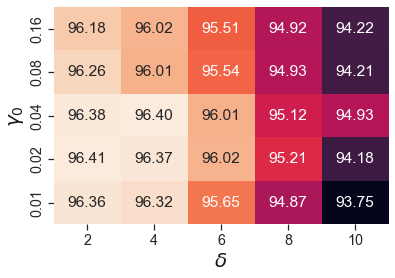}
	}
	\hfil
	\subfigure[The impact of decay factor $\delta$]{
		\includegraphics[width=0.47\columnwidth]{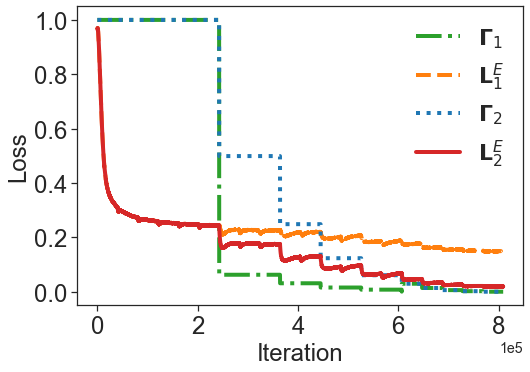}
	}
	\caption{Grid searching for the initial learning rate $\gamma_{0}$ and decay factor $\delta$. In (a), it shows the heatmap of $\gamma_{0}$ and $\delta$, where TAR@FAR$=1e-4$ on IJB-C is reported as the accuracy ($\%$). In (b), it illustrates the impact of the decay factor $\delta$, where $\delta=8$ for $\mathbf{\Gamma}_{1}$ and $\delta=2$ for $\mathbf{\Gamma}_{2}$. Thus, $\mathbf{\Gamma}_{1}$ declines fast and $\mathbf{L}^{E}_{1}$ goes stationary early}
	\label{fg:lr_df}
\end{figure}

\paragraph{\textbf{Initial learning rate $\gamma_{0}$ and decay factor $\delta$}} In Fig.~\ref{fg:lr_df}, we conduct a grid search to determine the optimal initial learning rate $\gamma_{0}$ and decay factor $\delta$. All results are reported based on ResNet100 trained for $5$ epochs on MS1MV3 with $1\times$GPU, while the test set is IJB-C. We observe a decrease in accuracy as $\delta$ increases, indicating that a large $\delta$ is detrimental to large-scale FR training. This is because a large $\delta$ causes the learning rate $\gamma_{t}$ to decrease rapidly, leading the model to stop learning prematurely when $\gamma_{t}$ becomes too small, as shown in Fig.~\ref{fg:lr_df}(b). Additionally, we find that an initial learning rate of $\gamma_{0}=0.02$ achieves the highest accuracy, since a smaller $\gamma_{t}$ halts the learning process, while a larger $\gamma_{t}$ results in instability. Based on these observations, we set $\gamma_{0}=0.02$ and $\delta=2$ for this study.

\begin{figure}[!t]
	\centering
	\subfigure[The heatmap of $\lambda$ and $\tau$]{
		\includegraphics[width=0.47\columnwidth]{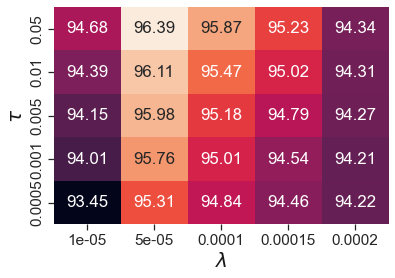}
	}
	\hfil
	\subfigure[Relationship of $\lambda$ and $\mathcal{D}^{E}_{t}$]{
		\includegraphics[width=0.47\columnwidth]{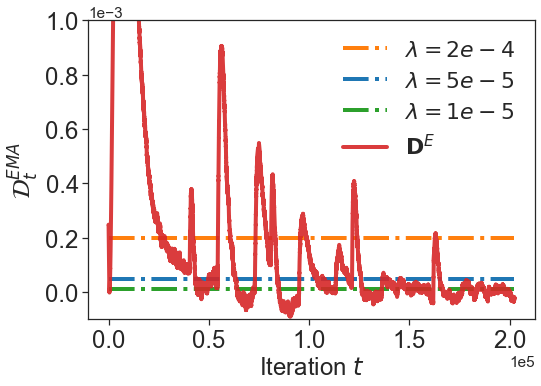}
	}
	\caption{Grid searching for the threshold $\lambda$ and tolerance $\tau$ in FastFace. In (a), it shows the heatmap of TAR@FAR$=1e-4$ on IJB-C is reported as the accuracy ($\%$). In (b), it illustrates the relationship of different $\lambda$ and $\mathcal{D}^{E}_{t}$. The curve of $\mathcal{D}^{E}_{t}$ under $\lambda$ will be treated as the signal to schedule the learning rate}
	\label{fg:thr_tol}
\end{figure}

\paragraph{\textbf{The threshold $\lambda$ and tolerance $\tau$ in FastFace}} In Fig.~\ref{fg:thr_tol}, we conduct a grid search to determine the optimal threshold $\lambda$ and tolerance $\tau$ in FastFace. All results are reported based on ResNet100 trained for $5$ epochs on MS1MV3 with $1\times$GPU, while the test set is IJB-C. We observe a decline in accuracy as $\tau$ decreases, indicating that a smaller $\tau$ is overly sensitive, which adjusts the learning rate prematurely. Conversely, a larger $\tau$ is too slow in detecting stationary subsequences. In addition, we find that $\lambda=5e-5$ achieves the highest accuracy. The reason is that, as shown in Fig.~\ref{fg:thr_tol}(b), a smaller $\lambda$ filters out most of $\mathcal{D}^{E}{t}$, while a larger $\lambda$ considers too many instances of $\mathcal{D}^{E}{t}$ as the signals to adjust the learning rate. Based on these observations, we set $\lambda=5e-5$ and $\tau=0.05$ for this study.

\paragraph{\textbf{The number of epochs $E$}} In Tab.~\ref{tab:epochs}, we compare the performance of PFC-0.3+FastFace across different training epochs. All results are reported based on ResNet100 on MS1MV3 with $1\times$GPU. A significant increase in accuracy is observed from $3$ to $5$ epochs, while the improvement slows down between $5$ and $7$ epochs. The increase from 3 to 5 epochs suggests that training for fewer than 5 epochs does not provide sufficient time for the model to converge. In addition, the slower increase from $5$ to $7$ epochs indicates that training beyond $5$ epochs contributes little to model convergence. As a consequence, we select 5 epochs as the optimal number for FastFace.

\begin{table}[!t]
	\centering
	\scalebox{0.85}{
		\begin{tabular}{lrrrrr}
			\toprule
			\textbf{Epoch} & \textbf{LFW} & \textbf{CFP-FP} & \textbf{AgeDB-30} & \textbf{IJB-B} & \textbf{IJB-C} \\
			\midrule
			3 & 99.50 & 98.08 & 97.83 & 93.43 & 94.76 \\
			4 & 99.65 & 98.46 & 98.01 & 94.80 & 96.03 \\
			5 & 99.80 & 98.53 & 98.12 & 95.03 & 96.41 \\
			6 & 99.80 & 98.55 & 98.11 & 95.06 & 96.55 \\
			7 & 99.83 & 98.53 & 98.12 & 95.08 & 96.65 \\
			\bottomrule
		\end{tabular}
	}
	\caption{The performance of PFC-0.3+FastFace on different training epochs. ResNet100 is selected as the model trained on MS1MV3. 1:1 verification accuracy ($\%$) is reported on LFW, CFP-FP, and AgeDB-30. TAR@FAR$=1e-4$ is reported as the accuracy ($\%$) on IJB-B and IJB-C}
	\label{tab:epochs}
\end{table}

\subsection{Ablation Study}
\paragraph{\textbf{The training time of FR methods with/without FastFace}} In Tab.~\ref{tab:ablation}, we compared the training time of ArcFace and PFC-0.3 on ResNet100 with/without FastFace. Note that without FastFace, the default Linear LR (LinLR) scheduler in PFC is selected, where the LR is linearly declining from $0.02$ to $0$. All approaches are trained on $1\times$GPU. Without FastFace, in order to ensure the convergence of the model, training for $20$ epochs is a common practice. It takes $116$H for ArcFace and $108$H for PFC-0.3, where the training time is reaching the limit of what we can endure. In addition, we also trained the model by ArcFace and PFC-0.3 with LinLR for $5$ epochs in order to ensure a more reasonable training time. However, it is observed a sharp drop in the accuracy on IJB-B and IJB-C. On the contrary, FastFace enables the model trained for $5$ epochs, while the accuracy is close to that for $20$ epochs. To be specific, $97.20$ compared to $97.51$. In summary, FastFace gives a competitive result with only $1/4$ of the original training time.

\begin{table}[!t]
	\centering
	\scalebox{0.85}{
		\begin{tabular}{lrrrrr}
			\toprule
			\textbf{Scheduler} & \textbf{Method} & \textbf{Epoch} & \textbf{Time} & \textbf{IJB-B} & \textbf{IJB-C} \\
			\midrule
			LinLR & ArcFace & 20 & 116H  & 95.22 & 97.31 \\
			LinLR & PFC-0.3 & 20 & 108H  & 95.87 & 97.51 \\
			LinLR & ArcFace & 5  & 35H   & 91.03 & 92.73 \\
			LinLR & PFC-0.3 & 5  & 30H   & 91.91 & 93.15 \\
			\midrule
			FastFace  & ArcFace & 5  & 36H   & 95.03 & 96.73 \\
			FastFace  & PFC-0.3 & 5  & 30H   & 95.68 & 97.20 \\
			\bottomrule
		\end{tabular}
	}
	\caption{The performance of ArcFace and PFC-0.3 with/without FastFace. ResNet100 is selected as the model and trained with $1\times$GPU. TAR@FAR$=1e-4$ is reported as the accuracy ($\%$) on IJB-B and IJB-C}
	\label{tab:ablation}
\end{table}

\paragraph{\textbf{FastFace across different ResNet models}} In Tab.~\ref{tab:model}, we present the training results of different models on MS1MV3 and WebFace12M. All models were trained for $5$ epochs on $1\times$GPU. Note the training on MV1MV3 utilized ArcFace with FastFace while that on WebFace12M employed CosFace with FastFace. We observed a progressive increase in training time from ResNet18 to ResNet200. Specifically, on MS1MV3, the training time increased from $4$ hours with a throughput of $1876$ images per second to $16$ hours with a throughput of $450$ images per second. On WebFace12M, the training time increased from $18$ hours with a throughput of $1015$ images per second to $30$ hours with a throughput of $590$ images per second. It is important to note that ResNet200 was not trained on WebFace12M because the computational demands exceeded the capabilities of the NVIDIA GeForce RTX 4090 GPU. A more significant increase in training time was observed for ResNet50 and ResNet100, indicating that the computational bottleneck shifts from other parts, e.g. data loading, in ResNet18, ResNet34, and ResNet50 to the model itself in ResNet100. For instance, on MS1MV3, the training times of ResNet18 and ResNet34 do not differ significantly when additional threads are not allocated for image reading, as the bottleneck is the speed of data loading. Additionally, the throughput on WebFace12M is lower than on MS1MV3 due to the greater number of identities in WebFace12M, which increases the computational time required for the FC layer in the model. Regarding accuracy, it is observed an increasing trend from ResNet18 to ResNet200 as we expected.

\begin{table}[!t]
	\centering
	\scalebox{0.85}{
		\begin{tabular}{lrrrrr}
			\toprule
			\textbf{Model} & \textbf{Dataset} & \textbf{images/s} & \textbf{Time} & \textbf{IJB-B} & \textbf{IJB-C} \\
			\midrule
			ResNet18  & MS1MV3     & 1876 & 4H  & 91.51 & 93.62 \\
			ResNet34  & MS1MV3     & 1423 & 5H  & 93.63 & 95.31 \\
			ResNet50  & MS1MV3     & 1203 & 6H  & 94.69 & 96.12 \\
			ResNet100 & MS1MV3     & 786  & 9H  & 95.03 & 96.41 \\
			ResNet200 & MS1MV3     & 450  & 16H & 95.16 & 96.65 \\
			\midrule
			ResNet18  & WebFace12M & 1015 & 18H & 92.13 & 94.43 \\
			ResNet34  & WebFace12M & 878  & 20H & 94.12 & 95.87 \\
			ResNet50  & WebFace12M & 783  & 23H & 95.25 & 96.85 \\
			ResNet100 & WebFace12M & 590  & 30H & 95.68 & 97.20 \\
			\bottomrule
		\end{tabular}
	}
	\caption{The performance of FastFace training different models. All models are trained for $5$ epochs with $1\times$GPU. TAR@FAR$=1e-4$ is reported as the accuracy ($\%$) on IJB-B and IJB-C}
	\label{tab:model}
\end{table}

\subsection{Comparison with Other LR Schedulers}

In Tab.~\ref{tab:lrs}, we compared FastFace with other commonly used LR schedulers. To be specific, we trained ResNet100 for $5$ epochs on Webface12M using CosFace, ArcFace, and PFC-0.3 with Linear LR Scheduler (LinLR), Multi-Step LR Scheduler (MSLR), Cosine Annealing LR Scheduler (CosLR), and FastFace. The LR in LinLR is linearly declining from $0.02$ to $0$ in total iterations $T$. MSLR halves the LR at $0.8T$, $0.6T$, $0.5T$, $0.4T$, and $0.2T$ respectively. CosLR schedules the LR by multiplying $\cos{\theta}$, where $\theta$ is increasing from $0$ to $\pi$ in the total iterations $T$. As shown in Tab.~\ref{tab:lrs}, FastFace outperforms the other three LR schedulers. The performance of CosLR is close to that of FastFace, as it provides a relatively smooth scheduling curve, starting with a large LR and gradually reducing it to a small value. This approach suits the practical need in deep learning for a fast learning pace initially, followed by slow fine-tuning in the later stages. In contrast, LinLR shows relatively lower accuracy compared to the other methods, including CosLR. This lower performance is due to the rapid reduction of the LR, which does not allow sufficient time for the model to make significant progress during the initial large LR phase, resulting in the need for extended training time due to the small LR.
Notably, PFC-0.3 with MSLR achieves an accuracy of $95.31$ on IJB-C, which is significantly higher than that of CosFace and ArcFace. This result is due to the hyper-parameters of MSLR being fine-tuned specifically for PFC-0.3, indicating that manually fine-tuning the LR scheduler can yield promising results. However, this process is time-consuming and impractical, much like manually fine-tuning the parameters in neural networks, which, while theoretically possible, is generally infeasible in practice due to the extensive time requirements.

\begin{table}[!t]
	\centering
	\scalebox{0.85}{
		\begin{tabular}{lrrrrr}
			\toprule
			\textbf{Scheduler} & \textbf{Method} & \textbf{LFW} & \textbf{CFP-FP} & \textbf{IJB-B} & \textbf{IJB-C} \\
			\midrule
			LinLR & CosFace & 99.63 & 98.46 & 91.23 & 92.93 \\
			LinLR & ArcFace & 99.65 & 98.73 & 91.03 & 92.73 \\
			LinLR & PFC-0.3 & 99.50 & 98.80 & 91.91 & 93.15 \\
			\midrule
			MSLR  & CosFace & 99.38 & 98.12 & 90.63 & 91.73 \\
			MSLR  & ArcFace & 99.50 & 98.27 & 90.93 & 92.05 \\
			MSLR  & PFC-0.3 & 99.80 & 99.09 & 94.79 & 95.31 \\
			\midrule
			CosLR & CosFace & 99.71 & 98.80 & 95.03 & 96.73 \\
			CosLR & ArcFace & 99.80 & 99.09 & 94.91 & 96.75 \\
			CosLR & PFC-0.3 & 99.80 & 99.11 & 95.22 & 97.09 \\
			\midrule
			FastFace  & CosFace & 99.81 & 99.11 & 95.46 & 97.09 \\
			FastFace  & ArcFace & 99.80 & 99.23 & 94.37 & 96.99 \\
			FastFace  & PFC-0.3 & 99.83 & 99.17 & 95.68 & 97.20 \\
			\bottomrule
		\end{tabular}
	}
	\caption{The performance of FastFace comparing with different LR schedulers on various FR methods. 1:1 verification accuracy is reported on LFW and CFP-FP, while TAR@FAR$=1e-4$ is reported on IJB-B and IJB-C}
	\label{tab:lrs}
\end{table}

\subsection{Comparison with SOTA Methods}

\begin{table*}[!t]
	\centering
	\scalebox{0.73}{
		\begin{tabular}{lrrrrrrrrr}
			\toprule
			\multirow{2}*{\textbf{Method}} & \multirow{2}*{\textbf{Dataset}} & \multirow{2}*{\textbf{\#GPUs}} & \multirow{2}*{\textbf{Epoch}} & \textbf{LFW} & \textbf{CFP-FP} & \textbf{AgeDB-30} & \textbf{IJB-B} & \textbf{IJB-C} \\
			\cmidrule(r){6-9}
			& & & & \multicolumn{3}{c}{$1:1$ Verification Accuracy} & \multicolumn{2}{c}{TAR@FAR=1e-4} \\
			\midrule
			CosFace~\cite{HWang:2018}        & Private               & 8          & 21 & 99.81 & 98.12  & 98.11 & 94.80 & 96.37 \\
			ArcFace~\cite{Deng1:2019}        & MS1MV2                & 4          & 16 & 99.83 & 98.27  & 98.28 & 94.25 & 96.03 \\
			CurricularFace~\cite{Huang:2020} & MS1MV2                & 4          & 24 & 99.80 & 98.37  & 98.32 & 94.80 & 96.10 \\
			Sub-center~\cite{Deng:2020}      & MS1MV3                & 8          & 24 & 99.80 & 98.80  & 98.31 & 94.94 & 96.28 \\
			MagFace~\cite{Meng:2021}         & MS1MV2                & 8          & 25 & 99.83 & 98.46  & 98.17 & 94.08 & 95.97 \\
			ElasticFace~\cite{Boutros:2022}  & MS1MV2                & 4          & 26 & 99.80 & 98.73  & 98.28 & 95.43 & 96.65 \\
			VPL~\cite{Deng:2021a}            & Cleaned MS1M          & 8          & 20 & 99.83 & 99.11  & 98.60 & 95.56 & 96.76 \\
			F$^{2}$C~\cite{Wang:2022}        & MS1MV2                & 8          & 20 & 99.50 & 98.46  & 97.83 & -     & 94.91 \\
			F$^{2}$C~\cite{Wang:2022}        & \textbf{WebFace42M}   & 8          & 20 & 99.83 & 99.33  & 98.33 & -     & 97.31 \\
			Virtual FC~\cite{PYLi:2021}      & Cleaned MS1M          & \textbf{1} & 16 & 99.38 & 95.55  & -     & 61.44 & 71.47 \\
			PFC-0.3~\cite{An:2022}           & WebFace4M             & 8          & 20 & 99.85 & 99.23  & 98.01 & 95.64 & 96.80 \\
			PFC-0.3~\cite{An:2022}           & WebFace12M            & 8          & 20 & \underline{99.83} & \underline{99.40}  & \underline{98.53} & \underline{96.31} & \underline{97.58} \\
			PFC-0.3~\cite{An:2022}           & \textbf{WebFace42M}   & 32         & 20 & \underline{99.85} & \underline{99.40}  & \underline{98.60} & \underline{96.47} & \underline{97.82} \\
			\midrule
			PFC-0.3+FastFace & MS1MV2     & \textbf{1} & \textbf{5} & 99.80 & 98.39  & 98.11 & 94.98 & 96.33 \\
			PFC-0.3+FastFace & MS1MV3     & \textbf{1} & \textbf{5} & 99.80 & 98.53  & 98.12 & 95.03 & 96.41 \\
			PFC-0.3+FastFace & WebFace4M  & \textbf{1} & \textbf{5} & 99.65 & 98.63  & 97.57 & 94.36 & 96.18 \\
			PFC-0.3+FastFace & WebFace8M  & \textbf{1} & \textbf{5} & 99.85 & 99.24  & 97.90 & 95.58 & 97.09 \\
			PFC-0.3+FastFace & WebFace12M & \textbf{1} & \textbf{5} & \underline{99.83} & \underline{99.17} & \underline{98.02} & \underline{95.68} & \underline{97.20} \\
			PFC-0.1+FastFace & \textbf{WebFace42M} & \textbf{1} & \textbf{5} & \underline{99.83} & \underline{99.33} & \underline{98.32} & \underline{96.01} & \underline{97.49} \\
			\bottomrule
		\end{tabular}
	}
	\caption{Performance comparisons between FastFace and other state-of-the-art FR methods on various benchmarks. For LFW, CFP-FP, and AgeDB-30, 1:1 verification accuracy ($\%$) is reported. For IJB-B and IJB-C, TAR@FAR$=1e-4$ is reported as the accuracy ($\%$).}
	\label{tab:compare}
\end{table*}

In Tab.~\ref{tab:compare}, we present a comparison of our method with other state-of-the-art (SOTA) approaches. Notably, while other methods require at least $4\times$GPUs and typically train for an average of $20$ epochs, FastFace achieves competitive results with just $1\times$GPU and only $5$ epochs. This efficiency translates to a significant reduction in training time, requiring only $1/4$ of the time compared to other methods. For example, training ResNet100 on WebFace12M using  PFC-0.3+FastFace takes just $30$ hours, as shown in Tab.~\ref{tab:model}. Despite the reduced resources and training time, FastFace’s performance still remains highly competitive. On WebFace12M, PFC-0.3+FastFace achieves an accuracy of $97.20$, which is close to the $97.58$ accuracy of PFC-0.3. Note that PFC-0.3 represents PFC-0.3 with Linear LR scheduler. Nevertheless, PFC-0.3 demands $8\times$GPUs and $20$ epochs, taking around $24$ hours for training. This comparison underscores the value of conducting large-scale training with just one GPU, as it shows that significant resource reductions can still yield strong performance.

In addition, we explored training ResNet100 using PFC-0.3 and F$^{2}$C on the much larger dataset, WebFace42M, which includes $42$M images with $2$M identities. While these methods achieve impressive accuracy on IJB-C, $97.82$ for PFC-0.3 and $97.31$ for F$^{2}$C, their training costs are substantial. For instance, PFC requires nearly $25$ hours with $32\times$GPUs, and F$^{2}$C takes days to complete with $8\times$GPUs. We also trained ResNet100 using PFC-0.1+FastFace on WebFace42M, reducing the batch size to 96. Note that the $0.1$ in PFC-0.1 means that randomly selecting $10\%$ class centers in each iteration. This training took around $6$ days, achieving an accuracy of $97.49$, just $0.33$ lower than PFC-0.3’s performance on WebFace42M. This finding aligns with the concept of diminishing marginal utility, where the cost increases significantly for minor improvements in FR model performance.

It is worth noting that Virtual FC also tries to train ResNet100 on self-cleaned MS1M with $1\times$GPU. Yet, it trains the model for $16$ epochs, which is considerably longer than the $5$ epochs needed by FastFace. Furthermore, its accuracy on IJB-B and IJB-C has a large gap compared to other methods as reported in~\cite{An:2022}. In contrast, FastFace maintains competitiveness on MS1MV2 and MS1MV3, reinforcing the effectiveness of our approach.

\section{Conclusion}

In this paper, we propose a fast-converging scheduler, named FastFace, for large-scale face recognition training. FastFace schedules the learning rate promptly and accurately in order to reduce the converging time of FR models. As a result, FastFace is able to train the model with $1/4$ of its original training time on $1\times$GPU by sacrificing less than $1\%$ accuracy. We conclude that large-scale FR training is now subject to the law of diminishing marginal utility, where costs increase rapidly for marginal improvements in model performance.

\paragraph{\textbf{Potential Societal Impacts}} FastFace enables large-scale FR training with one single GPU, making it accessible to researchers without extensive computing resources. By lowering the barrier to entry, this approach has the potential to stimulate growth and innovation in this research field.

\paragraph{\textbf{Limitations}} At the moment, FastFace contains two hyper-parameters, i.e. threshold $\lambda$ and tolerance $\tau$. In the future, we aim at eliminating these hyper-parameters entirely from FastFace. Additionally, we plan to conduct experiments on the Masked Face Recognition (MFR) challenge~\cite{Deng:2021b}.

\section*{Acknowledgment}
This work was supported in part by the Guangdong Basic and Applied Basic Research Foundation under Grant 2022A515110020, the Jinan University "Da Xiansheng" Training Program under Grant YDXS2409, the National Natural Science Foundation of China under Grant 62271232, and the National Natural Science Foundation of China under Grant 62106085.




\bibliographystyle{elsarticle-harv} 
\bibliography{pr}

\end{document}